\documentclass{article}

\usepackage{microtype}
\usepackage{graphicx}
\usepackage{subcaption}
\usepackage{booktabs} 

\usepackage{hyperref}

\usepackage{amsmath,amsfonts,bm}

\def\eqref#1{equation~\ref{#1}}

\def\1{\bm{1}}

\DeclareMathAlphabet{\mathsfit}{\encodingdefault}{\sfdefault}{m}{sl}
\SetMathAlphabet{\mathsfit}{bold}{\encodingdefault}{\sfdefault}{bx}{n}

\newcommand{\R}{\mathbb{R}}

\usepackage{url}
\usepackage{graphicx}
\usepackage{amsthm}
\usepackage{multirow}
\usepackage{adjustbox}
\usepackage{bm}
\usepackage[table]{xcolor} 
\usepackage{siunitx}
\usepackage{tikz}
\usetikzlibrary{positioning}
\theoremstyle{definition}

\newtheorem{lemma}{Lemma}[section]

\usepackage{comment}

\newcommand{\fl}{\operatorname{fl}}

\usepackage[preprint]{icml2026}

\usepackage{amsmath}
\usepackage{amssymb}
\usepackage{mathtools}
\usepackage{amsthm}

\usepackage[capitalize,noabbrev]{cleveref}

\theoremstyle{plain}

\theoremstyle{definition}

\theoremstyle{remark}

\icmltitlerunning{Induced Numerical Instability: Hidden Costs in Multimodal Large Language Models}

\begin{document}

\twocolumn[
  \icmltitle{Induced Numerical Instability: \\
    Hidden Costs in Multimodal Large Language Models}



  \icmlsetsymbol{equal}{*}

  \begin{icmlauthorlist}
    \icmlauthor{Wai Tuck Wong}{smu}
    \icmlauthor{Jun Sun}{smu}
    \icmlauthor{Arunesh Sinha}{rutgers}
  \end{icmlauthorlist}

  \icmlaffiliation{smu}{School of Computing and Information Systems, Singapore Management University, Singapore}
  \icmlaffiliation{rutgers}{Management Science and Information Systems Department, Rutgers Business School, New Jersey, USA}
  \icmlcorrespondingauthor{Wai Tuck Wong}{ wt.wong.2022@phdcs.smu.edu.sg}
  \icmlcorrespondingauthor{Sun Jun}{junsun@smu.edu.sg}
  \icmlcorrespondingauthor{Arunesh Sinha}{arunesh.sinha@rutgers.edu}

  \icmlkeywords{Machine Learning, ICML}

  \vskip 0.3in
]



\printAffiliationsAndNotice{}  

\begin{abstract}

The use of multimodal large language models has become widespread, and as such the study of these models and their failure points has become of utmost importance. We study a novel mode of failure that causes degradation in performance indirectly by optimizing a loss term that seeks to maximize numerical instability in the inference stage of these models. We apply this loss term as the optimization target to construct images that, when used on multimodal large language models, cause significant degradation in the output. We validate our hypothesis on state of the art models large vision language models (LLaVa-v1.5-7B, Idefics3-8B, SmolVLM-2B-Instruct) against standard datasets (Flickr30k, MMVet, TextVQA, VQAv2, POPE, COCO) and show that performance degrades significantly, even with a very small change to the input image, compared to baselines. Our results uncover a fundamentally different vector of performance degradation, highlighting a failure mode not captured by adversarial perturbations.
\end{abstract}

\section{Introduction}
\label{sec:intro}
Large language models (LLMs) have seen widespread deployment in recent years in many practical domains, from interfacing with customers through chatbots~\citep{DBLP:journals/corr/abs-2407-07858}, solving problems in healthcare~\citep{DBLP:journals/corr/abs-2505-08775},  to improving productivity in software engineering~\citep{DBLP:journals/corr/abs-2506-11060,DBLP:journals/corr/abs-2502-12115}.  These large language models are further equipped with tools (connected via the Model Context Protocol) to allow them to carry out tasks in the real world \cite{DBLP:journals/corr/abs-2508-14704}, allowing them to solve problems end-to-end completely autonomously as agents \citep{DBLP:journals/corr/abs-2505-20286}, often carrying out sensitive and critical tasks in the real world, such as governance in mission-critical systems \citep{DBLP:conf/icse-seip/EspositoPLT25}.

As their name suggests, LLMs rely on storing and loading a large number of parameters in memory on the order of billions to hundreds of billions \citep{DBLP:journals/tmlr/WeiTBRZBYBZMCHVLDF22}. This is often computationally and memory intensive \citep{DBLP:journals/corr/abs-2504-05299}. Using the standard float precision and without further optimizations, standard consumer GPUs can only store a model with about 6 billion parameters in memory (at 4 bytes per parameter, assuming 24GiB of VRAM). An optimization utilized in these models is to work with \textit{half precision}, at 2 bytes per parameter, effectively reducing the VRAM requirements by half. Further motivating this is the fact that computation is often much faster, as it can exploit hardware with SIMD (Single Instruction, Multiple Data) capabilities, which can process twice as many numbers simultaneously when using half-precision, leading to higher throughput \citep{DBLP:conf/iclr/MicikeviciusNAD18}. 

\begin{figure}[t]
\includegraphics[width=0.475\linewidth]{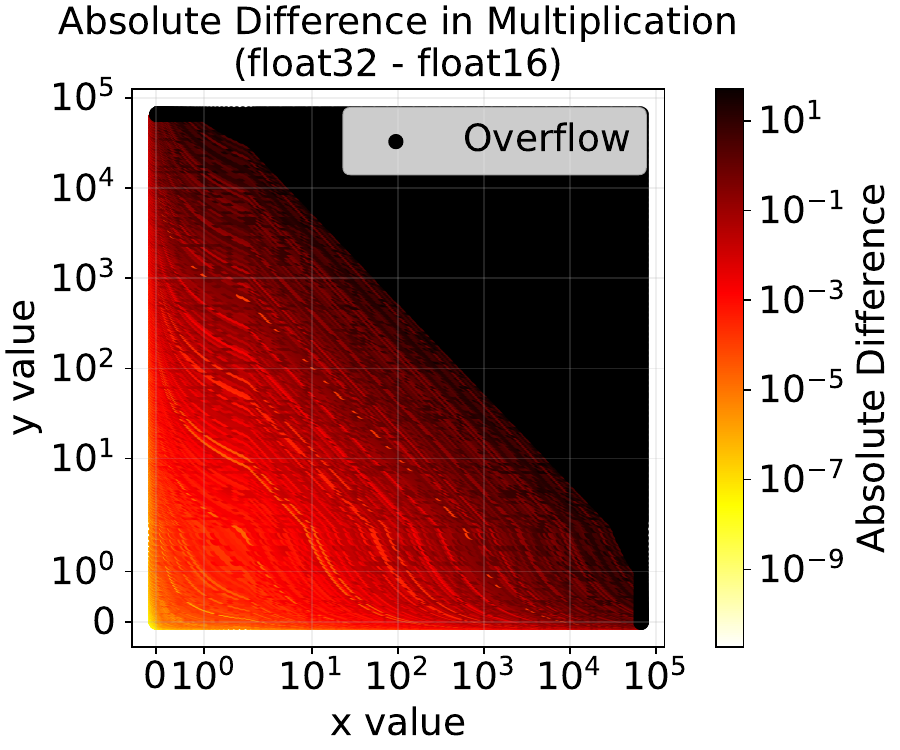}
\includegraphics[width=0.475\linewidth]{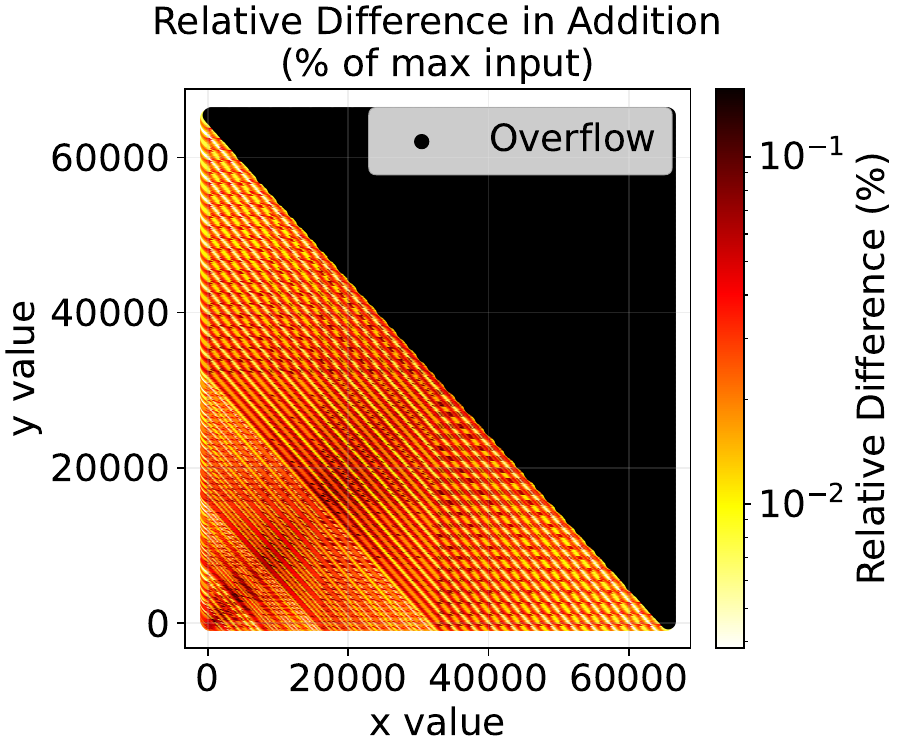}
\caption{We sample 300 input pairs, drawn uniformly from logarithmic space for multiplication and from linear space for addition, respectively, starting from the minimum and to the maximum positive values of a half-precision ($\mathrm{float16}$) number. We compare the absolute differences when we perform operations with both full and half precision as the data type We see that numerical imprecision generally increases as values increase for multiplication, meaning larger inputs values yield more numerical errors.}
\label{fig:abs-mutl-diff}

\end{figure}

This leads us to our first notion of instability. When a real number cannot be represented exactly in floating-point, it is rounded to the nearest representable value. At the \textbf{implementation level}, this means that the data type of the tensors results in rounding errors as operations are performed with finite precision, contributing to errors in the final output. Although the imprecision brought about by this optimization is often tolerable, this imprecision can, in fact,  cause unintended behavior in the underlying systems. We motivate this example with a simple multilayer perceptron (MLP) $y=Wx+c$. Indeed, even for such a simple MLP, when we observe the simple primitive operations of addition and multiplication (Fig. \ref{fig:abs-mutl-diff}) of half-precision numbers, these operations can lead to imprecision in the output. 

\begin{figure}[t]
  \centering
  \includegraphics[width=0.55\linewidth]{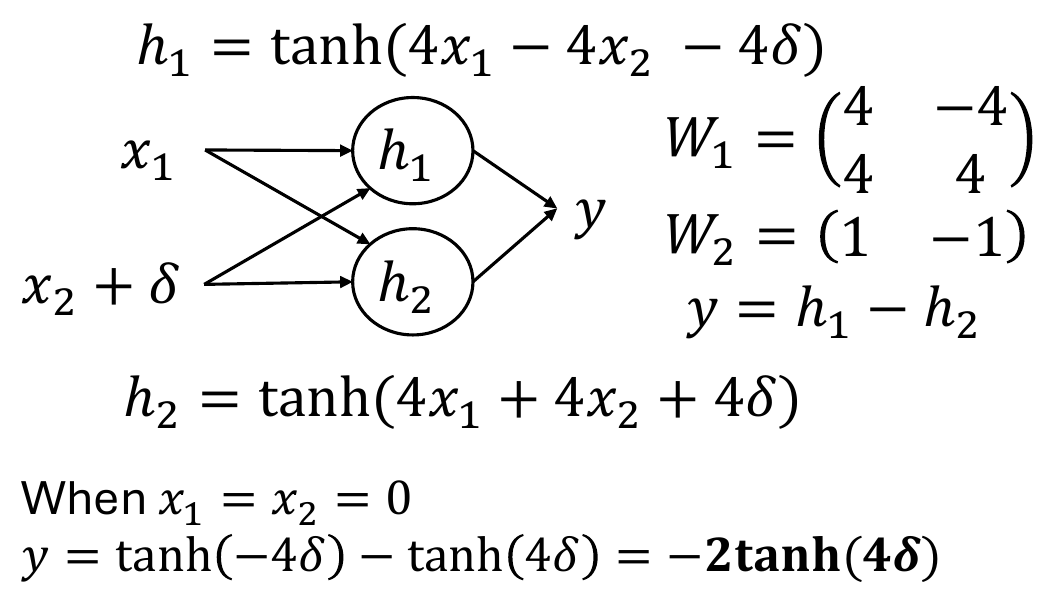}
  \includegraphics[width=0.40\linewidth]{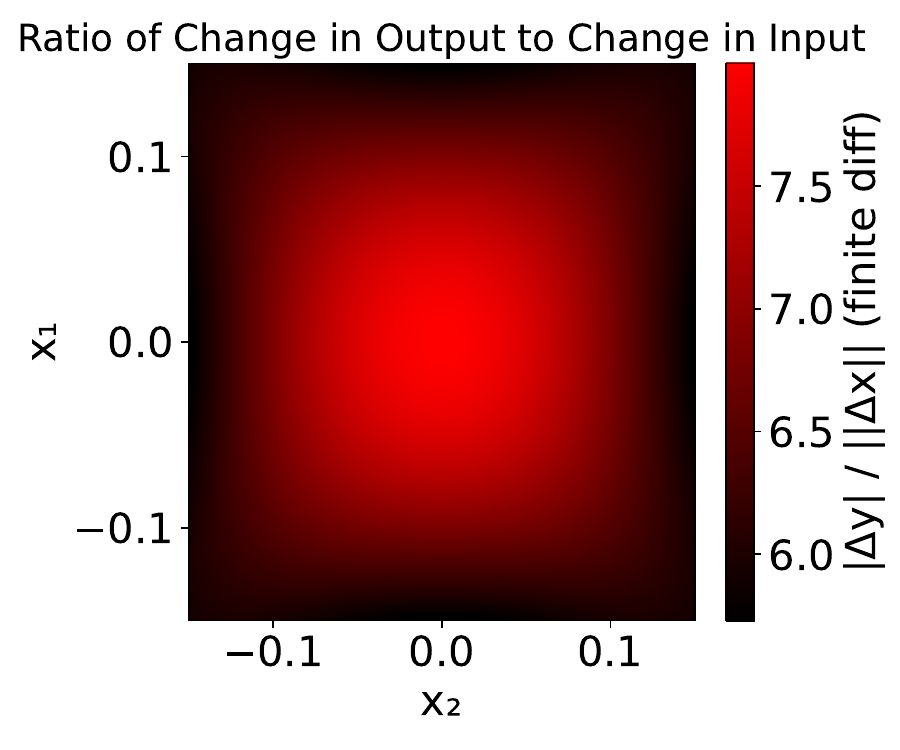}
  \caption{
    A two-layer $\tanh$ network illustrating how a small input perturbation $\delta$ 
    produces amplified change at the output due to asymmetric activation saturation. 
    When $x_1 = x_2 = 0$ and $x_2$ is perturbed slightly by $\delta$, the output becomes $y = - 2\tanh(4\delta)$. In this example the output around this region is around 8x the initial magnitude of the input perturbation,
    demonstrating local sensitivity at that region that is inherent within the network that cannot be resolved by increasing floating point precision.
  }
  \label{fig:tanh_instability}
\end{figure}

Our second notion of instability comes at the \textbf{function level}, where small changes in the input to the model can have large changes in the output. These errors are inherent within the parameters of the trained model and the operations in the model itself and, more importantly, cannot be solved by increasing the precision of the underlying data types (see the example in Fig. \ref{fig:tanh_instability}). We note that these errors constitute a distinct class of performance degradation and are different from adversarial perturbations, where in traditional adversarial attacks, we seek to maximize the \textbf{loss} of a model directly, whereas the approach of targeting numerical instability maximizes the \textbf{magnitude change in output} of a model. This effect is further exacerbated in multimodal large language models, such as large vision language models (LVLMs), where the visual input (an image) is typically projected into a latent space that is compatible with the text transformer. These soft embeddings are continuous in nature and can be sensitive to numerical perturbations, hence small changes in the image input may lead to unexpected errors downstream.

In this work, we leverage this insight and evaluate how numerical instability can affect performance of LVLMs. Specifically, we construct a method that induces numerical instability caused at both the implementation and functional level just from the input by modifying the image used in a text and image prompt when used for inference in a LVLM. We propose a novel computationally efficient loss function that serves as a proxy to the actual numerical instability, and optimize it via a whitebox, gradient-based approach. We show that the numerical imprecision caused by this modification causes significant degradation in performance in a wide range of tasks, specifically on standard image captioning and visual question and answer (VQA) datasets, in a manner that is different from traditional adversarial perturbations.

\section{Related Work}

\subsection{On Robustness in LVLMs}

Recent studies have explored the robustness of LVLMs, inspired by their leading performance on visual question answering tasks~\citep{DBLP:conf/nips/LiuLWL23a}. Current evaluations for robustness of LVLMs show that even simple Gaussian noise patches \citep{DBLP:journals/corr/abs-2504-01308} can cause degraded performance in LVLMs, and other more advanced methods of evaluation of robustness are derived from techniques used in adversarial attacks on machine learning models \citep{DBLP:journals/corr/GoodfellowSS14} which looks at maximizing the classification or task-based loss for a perturbed input. The techniques used for white-box attacks vary and include gradient-based attacks on the visual encoder portion of the LVLM \citep{DBLP:journals/corr/abs-2505-17440} and gradient-based approaches using APGD \cite{DBLP:conf/icml/Croce020a,DBLP:conf/cvpr/CuiAJL24}.

\begin{figure*}[t]
    \centering
    \begin{subfigure}{0.24\textwidth}
        \centering
        \includegraphics[width=\linewidth]{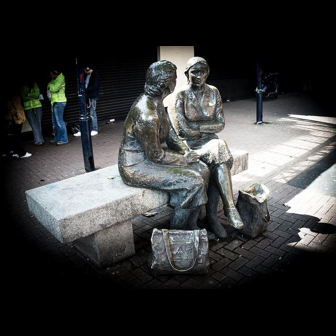}
    \end{subfigure}
    \begin{subfigure}{0.24\textwidth}
        \centering
        \includegraphics[width=\linewidth]{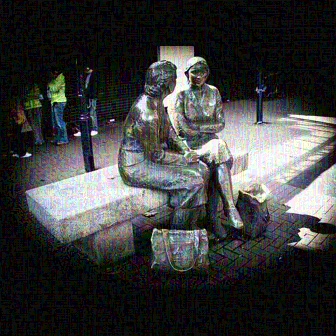}
    \end{subfigure}
    \begin{subfigure}{0.24\textwidth}
        \centering
        \includegraphics[width=\linewidth]{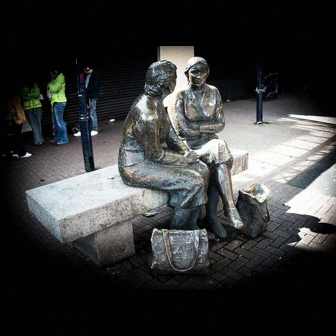}
    \end{subfigure}
    \begin{subfigure}{0.24\textwidth}
        \centering
        \includegraphics[width=\linewidth]{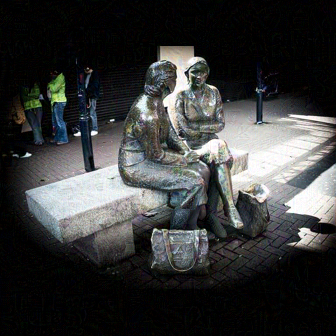}
    \end{subfigure}
    
    \begin{subfigure}{0.24\textwidth}
        \centering
        \includegraphics[width=\linewidth]{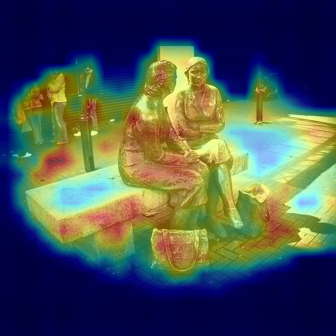}
        \caption*{\textbf{Clean Image}}
    \end{subfigure}
    \begin{subfigure}{0.24\textwidth}
        \centering
        \includegraphics[width=\linewidth]{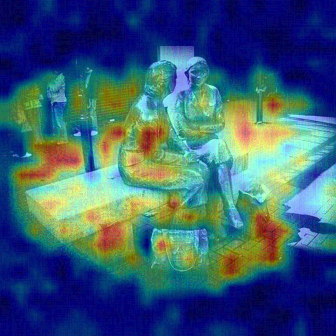}
        \caption*{\textbf{FGSM}}
    \end{subfigure}
    \begin{subfigure}{0.24\textwidth}
        \centering
        \includegraphics[width=\linewidth]{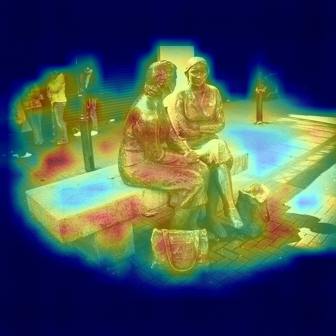}
        \caption*{\textbf{PGD}}
    \end{subfigure}
    \begin{subfigure}{0.24\textwidth}
        \centering
        \includegraphics[width=\linewidth]{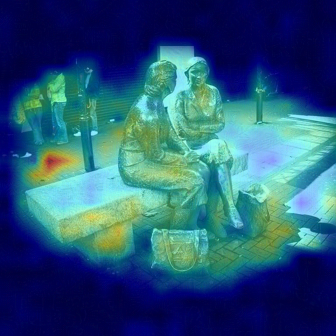}
        \caption*{\textbf{Ours (numerically unstable)}}
    \end{subfigure}

    \caption{\textit{Qualitative comparison of different perturbation types and their corresponding activation responses.}
The first row shows input images under four settings—clean, FGSM, PGD, and \textbf{numerically unstable (ours)}. 
The second row presents activation maps extracted from the first vision encoder layer of LLaVA-v1.5-7B. 
While adversarial perturbations (FGSM, PGD) introduce localized distortions, our numerically unstable input induces diffuse and misaligned attention, indicating degradation through a fundamentally different mechanism.}
    \label{fig:qualitative_results}
\end{figure*}

Our evaluation here is orthogonal to the attack techniques in literature; in particular, while these approaches aim to adversarially target the performance on a specific task (classification loss as an example), our goal is not task based, but inherent within the system itself, resulting in vastly different perturbations meant to \emph{induce numerical errors to affect task performance generically} (see Fig. \ref{fig:qualitative_results}).

Lipschitz-constrained networks have been explored in adversarial learning to bound sensitivity to input perturbations via spectral normalization, orthogonal (Parseval) regularization, or gradient penalties. While such constraints may appear applicable to mitigating instability in LVLMs, they do not scale well. Existing methods typically enforce per-layer Lipschitz bounds, as computing or constraining a global constant is intractable~\citep{zuhlke2025adversarial,li2023sok}. This layer-wise approach, while tractable for small CNNs or moderate-sized transformers, fails with the large depth of LVLMs, where hundreds of transformer blocks and attention maps make norm control expensive and unstable since the composition of many Lipschitz-bounded layers leads to either vanishing or exploding overall Lipschitz constants. Moreover, strict Lipschitz regularization tends to limit expressivity, conflicting with LVLMs’ need for sharp attention dynamics and nuanced semantic reasoning in the token space. Consequently, current LVLM architectures rely instead on implicit smoothness mechanisms, such as gradient clipping, normalization layers, rather than explicit Lipschitz enforcement.

\subsection{On Numerical Instability in Machine Learning}

It is widely known that numerical instability has negative effects in the lens of machine learning. Instability has been attributed for potentially explaining poor generalization and and training performance --- absorption errors for example has been known to cause the phenomenon of ``grokking'' where models do not improve for a very long time during training \citep{DBLP:conf/iclr/PrietoBMB25}. More notably, there has been recent interest in studying how instability can be coerced in mathematical operations directly to cause failure during the inference stage. Recent work has looked at how optimization layers can be targeted through the condition number of the underlying quadratic program to cause failure during inference \cite{DBLP:conf/aaai/WongKKNS23}. However, this evaluation is limited in scope and does not generalize well to other neural networks, including large language models. Similarly, other work \cite{DBLP:conf/sigsoft/YanCZTWW21} have looked at the failure conditions in vulnerable operations (such as division, exponential, etc.) in deep learning programs/models and uses back-propagation to find inputs to these models that would trigger these failure conditions. However, the identification of vulnerable operations requires static code analysis of the deep learning program, and the failure conditions are often fatal errors (e.g., divide by zero errors) and are not nuanced to the task at hand. Further, the operations targeted are not generally applicable to LLMs, which generally use much simpler elementary operations.

Our work is distinct in the following ways: the numerical errors we are studying is especially pertinent considering recent trends of reducing numerical precision to trade off memory efficiency; further, we propose a framework to define the loss in a way that does not need the ground truth but abuses properties inherent within the system, allowing us to generate perturbations that modify the image so that task performance drops during inference on the perturbed image. To the best of our knowledge, this is the first paper that applies the lens of studying numerical instability to understand the degradation of performance in multimodal large language models, specifically LVLMs.

\section{Methodology}

\subsection{Problem Statement}

The goal is to cause performance degradation at \textbf{inference time} on a LVLM $\mathbb{M}$, where the input provided to the model is the tuple $\mathbf{X} := (X^I, X^P) $ where $X^I$ is an image and $X^P$ is the text prompt. We evaluate numerical errors caused within some $\delta$  bounded by $\varepsilon$ around the image $X^I$ such that $\lVert \delta \rVert_\infty < \varepsilon$ over the image. As the prompt $X^P$ would be mapped to embeddings, the propagation of errors would be much more limited by the embedding space, and hence is omitted from our evaluation framework.

More concretely, the goal is to construct an error-maximizing input $\mathbf{X'} := (X'^I, X^P) $ where $X'^I:=X^I + \delta$. This would map to the worst case input causing the maximum numerical error for a given model $\mathbb{M}$. To facilitate the construction of the image, we backpropagate gradients when an image and text is passed through model $\mathbb{M}$ to the image, allowing us to update the image in the direction that maximizes the error.

\subsection{Methodology Details}

We first break down the forward pass of the model into its constituent \emph{elementary operations} as per \cite{DBLP:conf/nips/VirmauxS18}. Let the function $f$ be a function representing the model  $\mathbb{M}$, where $f: \mathbb{R}^n \rightarrow \mathbb{R}^m$ is computable in $K$ elementary operations, where $\exists (\theta_1,...,\theta_K)$ functions of the input $X$ and $(g_1,...,g_K)$ where $g_k$ is a real valued function of $(\theta_i)_{i\leq k-1}$ such that
\begin{equation}
\begin{aligned}
    &\theta_0(X)=X, (\theta_K(X), \ldots, \theta_{K-m+1}(X)) =f(X), \\ 
    & \forall k\in [1,K],\theta_k(X)=g_k(X,\theta_1(X),...,\theta_{k-1}(X))
\end{aligned}
\end{equation}

Intuitively, all operations $g_i$ in a model takes as input some prior intermediate results (represented by $\theta_i(X), i <k$) and the initial input $X$; the above represents the computational graph at the level of scalar computations.

Denote that hat operator $\hat\theta_k(X)_\mathbb{D} \in \mathbb{D}^{|\theta_k(X)|}$ as the representation of output of $\theta_k$ as a floating point number of limited precision, where $\mathbb{D}$ is the domain of all representable numbers in the system. Let $\theta_k(X) \in \mathbb{R}^{|\theta_k(X)|}$ correspond to $\hat{\theta}_k(X)$ with infinite precision. We define the numerical error due to finite precision as the absolute difference $E(g_k, \mathbb{D}) $ in the evaluation of the function $g_k$ on the infinite precision input $(X,\theta_1(X),...,\theta_{k-1}(X))$ viz-a-viz the limited precision $(\hat{X},\hat\theta_1(X),...,\hat\theta_{k-1}(X))_\mathbb{D}$ as follows:

\begin{equation}
    \begin{aligned}
        E(g_k, \mathbb{D}) 
        &= \vert g_k((X,\theta_1(X),...,\theta_{k-1}(X))) \\ &- g_k((\hat{X},\hat\theta_1(X),...,\hat\theta_{k-1}(X))_\mathbb{D}) \vert \\
        &= \vert \theta_k(X) - \hat\theta_k(X)_\mathbb{D} \vert
    \end{aligned}
\end{equation}

In practice, $g_k$ can be any elementary operation, from addition, subtraction, multiplication, division, matrix multiplication, to softmax and even the LayerNorm. Given an input $X'^I = X^I + \delta$ to the neural network, for any intermediate operation $\theta_k(X'^I)=g_k(X'^I,\theta_1(X'^I),...,\theta_{k-1}(X'^I))$, the input to the function are floating point tensors $(X'^I,\hat\theta_1(X'^I),...,\hat\theta_{k-1}(X'^I)))_\mathbb{D}$, where each input tensor depends on the neural network input $X'^I = X^I + \delta$. It follows that we want to optimize the following objective to maximize the numerical error in all floating point operations performed in the model.

\begin{equation}
    \begin{aligned}
        & \max_\delta \sum_{k \in[1,K]} E(g_k,\mathbb{D}) \\ 
        = &\max_\delta \sum_{k \in[1,K]} \vert \theta_k(X'^I) - \hat\theta_k(X'^I)_\mathbb{D} \vert \\
        = &\max_\delta \sum_{k \in[1,K]} \vert \theta_k(X^I + \delta) - \hat\theta_k(X^I + \delta)_\mathbb{D} \vert
    \end{aligned}
\end{equation}

$$$$

However, this formulation is intractable, as calculating the infinite precision $\theta_k(X^I + \delta)$ for each and every operation is computationally intractable.
The mathematical results below motivate the design of a proxy loss, which we present subsequently.

We first describe the IEEE 754 floating-point standard~\citep{DBLP:books/daglib/0007208}, which rounds to the nearest representable number given the available precision. In cases where the number lies exactly midway between two representable values, the standard specifies the round half to even rule (also known as banker’s rounding), meaning that the value is rounded to whichever of the two candidates has an even least significant digit. Both CUDA and PyTorch use IEEE 754 style rounding for float32 and float64, and a slightly modified version for float16. In all cases, it is possible to represent the rounding of $x \in \R$ as 
$\fl(x)=x(1+k)$ with 
$|k|\le \epsilon$ where $\epsilon$ is a constant that depends on the float precision; $\epsilon$ is typically small with values $2^{-11}$ for float16, $2^{-24}$ float32 and so on~\citep{ieee754-2019}. Note that the rounding error $k x$ increases with the magnitude of $x$. This observation can be extended to the evaluation of the function $f$ on inputs and outputs restricted to a given precision.

\begin{lemma}[Forward error bound under floating-point input rounding and result rounding]
Let $f:\R^d\to\R^k$ be locally Lipschitz on an open set 
$U \subset \R^d$ with Lipschitz cosntant $L$. Assume IEEE 754 style rounding as stated above and generalized to multiple dimension as $(\fl(x))_j =x_j (1 + \eta_j )$ with 
$|\eta_j |\le \epsilon $ for $x = (x_1, \ldots, x_j, \ldots, x_d) \in \R^d$, and further assume function output rounding: $\widehat{y}_j= (\fl\big(f(\fl(x))\big))_j =f_j(\fl(x))(1+ \eta_j)$ with 
$| \eta_j |\le \epsilon$. Fix a compact set $K \subset U$ such that $x, \fl(x) \in K$. Then the absolute forward error satisfies
\begin{equation}
\|\widehat{y}-f(x)\|
\le
\underbrace{L\,\epsilon\|x\|}_{\text{propagated input rounding}}
+\,
\underbrace{\epsilon\,\|f(x)\|+O(\epsilon^2)}_{\text{result rounding}}
\label{eq:rounding_error_bound}
\end{equation}
where the $O(\epsilon^2)$ term is uniform on $U$ (i.e., does not depend on $x$).
\end{lemma}
\begin{proof}
Let $\Delta x = \fl(x) - x$ and $\Delta y = \widehat{ y} - f(\fl (x) )$. Then, it is straightforward to check that $\| \Delta x \|  \leq \epsilon \| x \|$ and $\Delta y \leq \epsilon \| f(\fl(x)) ||$.
Start with the triangle inequality to get
\begin{equation}
\begin{aligned}
\|\widehat{y} - f(x)\|
&= \|f(\fl(x)) + \Delta y - f(x)\| \\[2pt]
&\leq \|f(\fl(x)) - f(x)\| + \|\Delta y\| \; 
\end{aligned}
\end{equation}
By Lipschitz continuity, the first term on the RHS above can be written as
\begin{equation}
    \|f(\fl(x)) -f(x)\| \leq L \| \fl(x) -x \| \leq L \epsilon \| x \|
\end{equation}

For the second term,
\begin{equation}
    \begin{aligned}
   \|\Delta y\| \leq \epsilon \| f(\fl(x)) \| &\leq \epsilon (\| f(x) \| + \| f(\fl(x))  - f(x)\|) \\
   & \leq \epsilon (\| f(x) \| + L \epsilon^2 \| x \| 
\end{aligned}
\end{equation}

Combining this and noting the fact $\| x \| \leq M$ for some constant $M$ (since $K$ is compact), yields the result
\begin{equation}
\|\widehat{y}-f(x)\|
\;\le\;
L\,\epsilon\|x\|
\;+\;
\epsilon\,\|f(x)\| + O(\epsilon^2) \;
\label{eq:rounding_error}
\end{equation}
\end{proof}

Since the numerical errors increase with the magnitude of the input, with this result, we simply have to increase the magnitude of the input $(\vert \hat\theta_1(X'^I)\vert,...,\vert \hat\theta_{k-1}(X'^I) \vert)_\mathbb{D}$ in every operation $g_k$ to increase the numerical error. Consequently, we use the following proxy objective to maximize the numerical error, by forcing an increase in every input to every function $g_k$, that is,

\begin{align}
  \max_\delta \sum_{k \in[1,K]} E(g_k,\mathbb{D}) & \approx \max_\delta \sum_{k \in[1,K]} \vert \hat\theta_k(X'^I)_\mathbb{D} \vert \nonumber\\ 
  &\!\!\!\!\!\!\!\!\!\!\!\!\!\!\!\!\!\!\!\!\!\!= \max_\delta \sum_{k \in[1,K]} \vert \hat\theta_k(X^I + \delta)_\mathbb{D} \vert
\label{eq:approx_loss}
\end{align}


Although the proxy loss formulation accounts only for errors arising from finite precision, maximizing the magnitude of each input $\vert \hat\theta_k(X^I + \delta)_\mathbb{D} \vert$ introduces an implicit coupling effect: it drives substantial variations in the output of the preceding functions, which in turn becomes the input to $g_k$. In doing so, we \textbf{locally maximizing the change in output at every point in the model.}
More precisely, every input to the function $g_k$ is an output of some previous function evaluation $\theta_i(X^I + \delta)_\mathbb{D}$, for $i \leq k-1$. Then, suppose for a function $g_{k}$ preceded by functions $\theta_i$'s, the change in input of $g_{k}$ is $\gamma$ (propagated from the change of $\delta$ in input to the neural network). Let $Z$ be the unperturbed input to $g_k$. Then, the loss trying to maximize $\vert \theta_k(Z + \gamma) \vert$ when $\vert \theta_k(Z) \vert$ is small is as follows:

\begin{equation}
\arg\!\max_\gamma  \vert \theta_k(Z +\gamma) \vert \approx \arg\!\max_\gamma \vert \theta_k(Z+\gamma) - \theta_k(Z) \vert
\end{equation}


In other words, this loss function heuristically finds the $\delta$ that aims to cause a large change in $\theta_k(Z)$, particularly when $\theta_k(Z)$ is small, at every function evaluation in the network, i.e., the loss attempts to find a change in the input $\gamma$ (propagated from the change of $\delta$ in input $X$) within a local neighborhood that induces the largest change in the output at each primitive function. This goal may not be achieved for every function evaluation in the network but on aggregate achieving the largest possible change for some function evaluations will ultimately induce the network to behave in a numerically unstable way, not just from numerical precision but also at a functional sensitivity level.

\begin{figure*}[t]
    \centering

    \begin{subfigure}{0.23\textwidth}
        \centering
        \includegraphics[width=\linewidth]{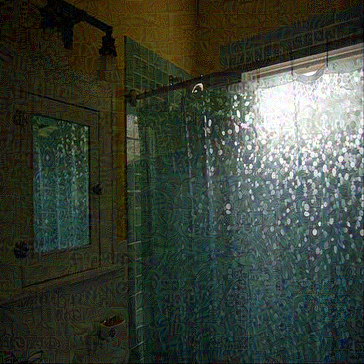}
        \vspace{1mm}
        \begin{minipage}[t][28mm][t]{\linewidth}  
            \scriptsize
            \textbf{Question:} What material is the walls made of?\\
            \textbf{Ground Truth:} Tile.\\
            \textbf{Clean Caption:} Tile.\\
            \textbf{Perturbed Caption:} Glass.
        \end{minipage}
    \end{subfigure}
    \hfill
    \begin{subfigure}{0.23\textwidth}
        \centering
        \includegraphics[width=\linewidth]{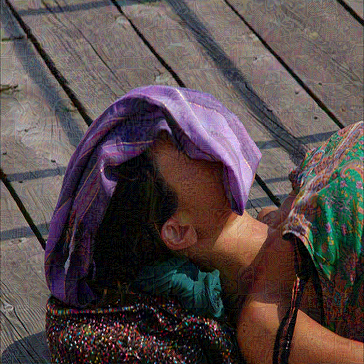}
        \vspace{1mm}
        \begin{minipage}[t][28mm][t]{\linewidth}
            \scriptsize
            \textbf{Question:} Write a short caption for this image.\\
            \textbf{Ground Truth:} A girl is sunning herself with a purple towel over her face.\\
            \textbf{Clean Caption:} A woman wearing a purple scarf lays on a wooden surface.\\
            \textbf{Perturbed Caption:} The purple shirt man is fighting with the other man.
        \end{minipage}
    \end{subfigure}
    \hfill
    \begin{subfigure}{0.23\textwidth}
        \centering
        \includegraphics[width=\linewidth]{sec/figures/example3.png}
        \vspace{1mm}
        \begin{minipage}[t][28mm][t]{\linewidth}
            \scriptsize
            \textbf{Question:} Write a short caption for this image.\\
            \textbf{Ground Truth:} There is a piece of cake on a plate with decorations on it.\\
            \textbf{Clean Caption:} The plate has a caramel design on top of the dessert.\\
            \textbf{Perturbed Caption:} A plate of food that looks like a steak with veggies on it.
        \end{minipage}
    \end{subfigure}
    \hfill
    \begin{subfigure}{0.23\textwidth}
        \centering
        \includegraphics[width=\linewidth]{sec/figures/example4.png}
        \vspace{1mm}
        \begin{minipage}[t][28mm][t]{\linewidth}
            \scriptsize
            \textbf{Question:} What town are we in?\\
            \textbf{Ground Truth:} Burnaby.\\
            \textbf{Clean Caption:} Burnaby.\\
            \textbf{Perturbed Caption:} Newark.
        \end{minipage}
    \end{subfigure}

    \vspace{-9mm}

    \caption{Examples of outputs under clean and numerically unstable inputs ($\varepsilon=16/255$) for the Idefics3-8B model. Each column shows the perturbed input image, the associated question, and corresponding responses. The perturbed responses deviate semantically from both the ground truth and the model’s original answers on the clean image, despite nearly identical inputs. Additional examples for other models can be found in Appendix \ref{sec:additional_attack_outputs}.}
    \label{fig:vqa_poisoned_examples}
\end{figure*}

\subsection{Implementation Details}

Maximizing the proxy loss developed above through gradient descent, however, proves challenging, as inducing numerical instability causes the optimization process to behave unexpectedly and fail to converge, either because of vanishing or inaccurate gradients. We do the following tricks to ensure that the optimization goes smoothly.

\textbf{Mixed Precision}: As the inputs that are passed to the model are of $\mathrm{float16}$, operations that aim to accumulate loss as per Eq. \ref{eq:approx_loss} are also prone to numerical imprecision and would lead to inaccurate gradients. Further, the $\delta$ that we are optimizing may also lose precision if stored and updated in $\mathrm{float16}$. As a workaround, we store the a ``master copy'' of the perturbations $\delta$ as a $\mathrm{float64}$ tensor, as well as any accumulated loss as $\mathrm{float64}$ tensors, to avoid losing precision leading to inaccurate gradients, and specifically when accumulating the loss, we upcast and use $\mathrm{float64}$.

\textbf{Optimization}: As the gradients were observed to be small in terms of magnitude during the optimization process, the update can potentially be numerically imprecise because the gradients themselves when backpropagated are $\mathrm{float16}$ tensors, leading to an unstable training regime. We use a method analogous to gradient scaling in \cite{DBLP:conf/iclr/MicikeviciusNAD18} by employing a similar approach to the iterative Fast Gradient Sign Method (FGSM) \citep{DBLP:conf/iclr/KurakinGB17a}. Through this, we circumvent the tiny magnitude of gradients at each iteration by scaling the update to the step size $\alpha$ using the sign of gradient rather than the actual small value, i.e., we perform the following update at each epoch:
\begin{equation}
    \delta'_{i+1} = \delta'_{i} + \alpha \cdot \text{sign} (\nabla_{\delta_i'}\text{loss})
\end{equation}

where the loss is as defined in Eqn \ref{eq:approx_loss}. Different from FGSM, we do not use the prediction loss in our induced numerical instability.


\section{Experiments}

\begin{table*}[h]
\centering
\renewcommand{\arraystretch}{1.25}
\setlength{\tabcolsep}{8pt}

\begin{adjustbox}{max width=\textwidth}
\begin{tabular}{llccccc}
\toprule
& & \multicolumn{2}{c}{\textbf{Image Captioning}} & \multicolumn{3}{c}{\textbf{Visual Question Answer}} \\
\cmidrule(lr){3-4} \cmidrule(lr){5-7}
Model & $\delta$ Type & MSCOCO & Flickr30k & TextVQA & VQAv2 & POPE \\
\midrule

\rowcolor{gray!20}
\multirow{4}{*}{\textbf{LLaVA-v1.5-7B}} 
& \textbf{NUM (ours)}  & $\bm{0.845\pm0.00}$ & $\bm{0.639\pm0.01}$ & $\bm{0.337\pm0.01}$ & $0.746\pm0.00$ & $\bm{0.840\pm0.00}$ \\
& RAND  & $1.13 \pm 0.01$ & $0.854\pm0.00$ & $0.438\pm0.00$ & $0.750\pm0.00$ &  $0.843\pm0.01$   \\
& GAUS  & $1.13\pm0.00$ & $0.862\pm0.00$ & $0.431\pm0.01$ & $\bm{0.741\pm0.00}$ & $0.848\pm0.01$ \\
& NONE  & $1.14\pm0.00$ & $0.869\pm0.00$ & $0.446\pm0.00$ & $0.747\pm0.00$ & $0.840\pm0.00$ \\
\midrule

\rowcolor{gray!20}
\multirow{4}{*}{\textbf{Idefics3-8B}} 
& \textbf{NUM (ours)} & $\bm{0.273\pm0.01}$ & $\bm{0.221\pm0.01}$ & $\bm{0.327\pm0.00}$ & $\bm{0.560\pm0.00}$ & $\bm{0.723\pm0.01}$ \\
& RAND  & $0.649\pm0.00$ & $0.587\pm0.01$ & $0.569\pm0.00$ & $0.772\pm0.00$ & $0.833\pm0.01$  \\
& GAUS  & $0.647\pm0.01$ & $0.601\pm0.01$ & $0.567\pm0.01$ & $0.770\pm0.01$ & $0.823\pm0.00$  \\
& NONE  & $0.664\pm0.00$ & $0.557\pm0.00$ & $0.593\pm0.00$ & $0.783\pm0.00$ & $0.840\pm0.00$  \\
\midrule

\rowcolor{gray!20}
\multirow{4}{*}{$\bm{\text{SmolVLM2-2.2B}}$} 
& \textbf{NUM (ours)} & $\bm{0.826\pm0.00}$ & $\bm{0.697\pm0.00}$ & $\bm{0.387\pm0.00}$ & $0.689\pm0.00$ & $\bm{0.774\pm0.00}$ \\
& \text{RAND} & $0.957\pm0.01$ & $0.761\pm0.01$ & $0.427\pm0.01$ & $\bm{0.666\pm0.00}$ & $0.783\pm0.00$ \\
& \text{GAUS} & $0.950\pm0.00$ & $0.760\pm0.00$ & $0.424\pm0.00$ & $0.673\pm0.00$ & $0.777\pm0.01$ \\
& \text{NONE} & $0.975\pm0.00$ & $0.759\pm0.00$ & $0.456\pm0.00$ & $0.701\pm0.00$ & $0.794\pm0.00$ \\
\midrule
\rowcolor{gray!20}
\multirow{4}{*}{$\bm{\text{Janus-Pro-1B}}$} 
& \textbf{NUM (ours)} & $\bm{0.454\pm0.00}$ & $\bm{0.412\pm0.00}$ & $\bm{0.361\pm0.01}$ & $0.612\pm0.00$ & $\bm{0.818\pm0.01}$ \\
& \text{RAND} & $0.506\pm0.00$ & $0.451\pm0.02$ & $0.464\pm0.00$ & $0.630\pm0.00$ & $0.857\pm0.00$ \\
& \text{GAUS} & $0.519\pm0.01$ & $0.457\pm0.01$ & $0.466\pm0.00$ & $0.635\pm0.00$ & $0.850\pm0.00$ \\
& \text{NONE} & $0.466\pm0.00$ & $0.415\pm0.00$ & $0.485\pm0.00$ & $\bm{0.608\pm0.00}$ & $0.866\pm0.00$ \\
\bottomrule
\end{tabular}
\end{adjustbox}

\caption{Evaluation of models under different perturbations and noise across image captioning and visual question answering benchmarks. For Image Captioning task, the lower the semantic similarity scores (CIDEr-D) compared to the ground truth, the greater the degradation in performance. For the Visual Question Answering task, the lower the accuracy (\%), the greater the degradation in performance. We highlight the perturbation type that caused the most degradation in \textbf{bold}.}
\label{table:main_result}
\end{table*}

\subsection{Experimental Setup}

The goal of our evaluation is to understand how numerical instability affects LVLM performance under different conditions when induced by the optimization of Eq. \ref{eq:approx_loss} for an input image and to show that this applies across a diverse range of tasks and models of various architectures and sizes.

For our evaluation, we choose a range of representative pretrained state-of-the-art vision large language models with varying number of parameters, namely LLaVa-v1.5-7B \citep{DBLP:conf/cvpr/LiuLLL24}, Idefics3-8B \citep{DBLP:journals/corr/abs-2408-12637}, SmolVLM-2.2B-Instruct \citep{DBLP:journals/corr/abs-2504-05299} and Janus-Pro-1B \citep{DBLP:journals/corr/abs-2501-17811}, and we evaluate them on standard multimodal tasks as described below. Specifically, for each of these tasks, given an input image and prompt $\mathbf{X} := (X^I, X^P) $, we aim to find a $\delta, \vert \delta \vert < \varepsilon$ that maximizes the loss function in Eq. \ref{eq:approx_loss}, with $\varepsilon = 16/255$ which is often adopted in existing studies \citep{DBLP:journals/corr/abs-2505-17440}. All instability-inducing experiments are conducted over $t=100$ iteration steps. All optimizations are performed using the AdamW optimizer \citep{DBLP:conf/iclr/LoshchilovH19} in PyTorch, with hyperparameters $\gamma=0.2$, $\alpha=0.01$. As image sizes vary for each of the following tasks, all images are either resized (if the model supports it) or padded and scaled down. Values are averaged over 3 runs with standard deviations reported. All experiments are conducted on either a NVIDIA H100 NVL or Nvidia H200 SXM5 GPU.

\textbf{Image Captioning}: For the image captioning task, the goal for the LVLM is to generate a caption that matches the ground-truth captions based on the image provided. We use the prompt ``Write a short caption for this image.'' to generate the caption for a given image. For this task, we evaluated the models on 500 validation images from the MSCOCO dataset \citep{DBLP:conf/eccv/LinMBHPRDZ14}, and 500 test images from the Flickr30k dataset \citep{DBLP:journals/tacl/YoungLHH14}. We evaluate the generated captions using CIDEr-D \citep{DBLP:conf/cvpr/VedantamZP15} as a measure of distance to determine the degradation in task performance against the ground truth captions.

\textbf{Visual Question \& Answering}: For the visual question \& answering task, given an image and a question, the goal for the LVLM is to provide a correct answer for the question based on the contents of the image. For the image, we use the image as is with the standard preprocessing, and for the prompt we provide the original question and append an additional line ``Answer in as few words as possible, giving only the answer.'' We evaluate the models using standard datasets in this domain, namely 500 images from the validation set of TextVQA~\citep{DBLP:conf/cvpr/SinghNSJCBPR19}, 500 images from the validation set of VQAv2~\citep{DBLP:journals/ijcv/GoyalKASBP19}. Additionally, similar to other work, we further evaluate the impact on inference of the models using hallucination benchmarks like POPE~\citep{DBLP:conf/emnlp/LiDZWZW23}. We use the standard VQA accuracy~\citep{DBLP:conf/iccv/AntolALMBZP15} as a measure of degradation in task performance.

\textbf{Baselines}  We compare the performance degradation induced by our induced numerical error method (\textbf{NUM}) against several baselines: no noise (\textbf{NONE}), uniformly sampled random noise in the range $[-\varepsilon, \varepsilon]$ (\textbf{RAND}), and Gaussian noise with mean 0 and standard deviation 0.1 bounded to $\varepsilon$ (\textbf{GAUS}), following \cite{DBLP:journals/corr/abs-2504-01308}. Because this represents a distinct and previously unexamined failure mode, such comparisons highlight its qualitative difference from other types of noise.

\subsection{Main Results}

\begin{figure}[t]
\centering
\includegraphics[width=0.8\linewidth]{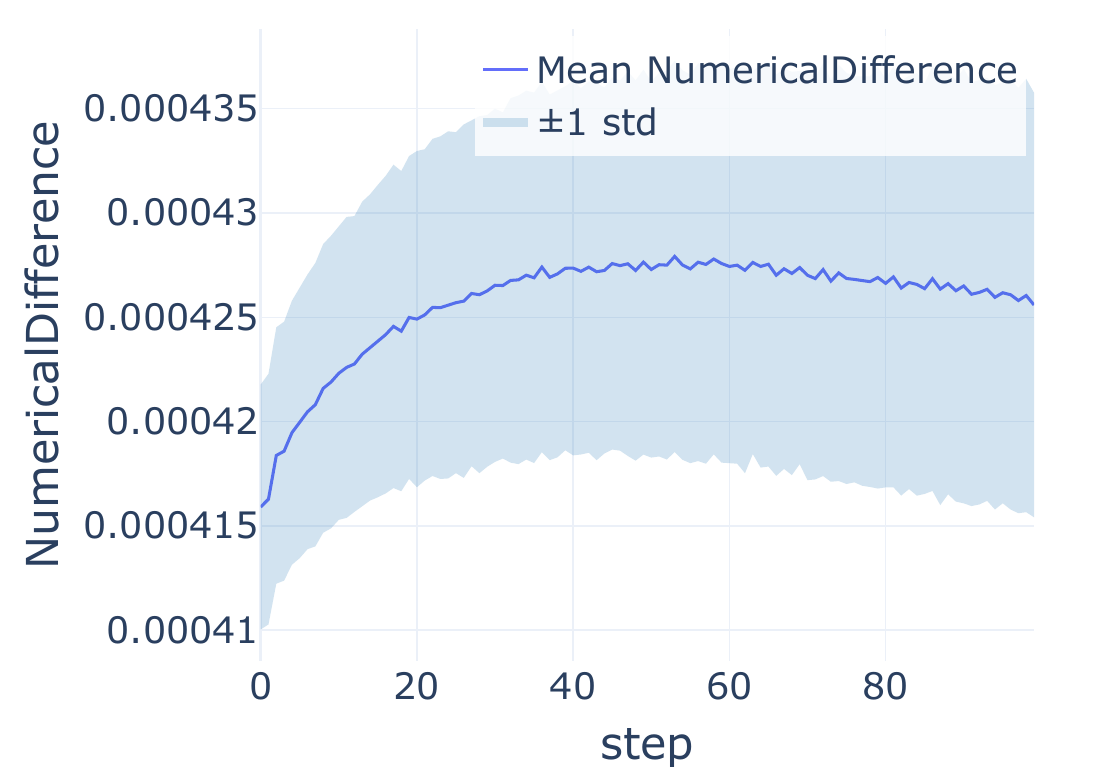}
\caption{Implementation-level numerical error on MSCOCO for Idefics3-8B. We plot accumulated absolute differences between $\mathrm{float32}$ and $\mathrm{float16}$ forward passes (summed over operations and samples) versus training epoch (optimizing Eqn.~\ref{eq:approx_loss}). Numerical error increases over epochs under our proxy loss.}
\label{fig:num-diff-idefics}

\end{figure}

We evaluate the impact of our method against baselines with the same perturbation budget of $\varepsilon = 16/255$ (see Appendix~\ref{ablation} for additional results for other $\varepsilon$) on various tasks for LVLMs and report the results in Table \ref{table:main_result}. We see that optimizing for the proxy loss in Eqn. \ref{eq:approx_loss} corresponds to an accumulation of numerical error measured at the implementation level (see Fig. \ref{fig:num-diff-idefics}) and the degradation on task performance is also compounded by the function level numerical instability. Here, we see up to a 59\% drop in performance (for the Idefics3-8B model on MSCOCO, from 0.664 to 0.273) when applying our method \textbf{NUM}, and our method shows a stronger degradation in task performance compared to baseline perturbations, with the baselines showing a much lower maximum decrease at (5\% in the VQAv2 task for the SmolVLM2 model, from 0.701 to 0.666). 

However, we do note that the average performance decrease for some tasks appear to be smaller when using standard metrics, especially for the VQA family of tasks. Specifically, the reported VQA accuracy as used in generative models such as LVLMs may not be reflective of the true degradation of task performance of the model as the definition of the VQA metric does not accounting for synonyms (since the machine generated answer must be contained exactly within the list of correct answers after some pre-processing for it to contribute to accuracy). To better capture semantic alignment beyond exact matching, we compute the cosine similarity between Sentence-BERT embeddings~\citep{DBLP:conf/emnlp/ReimersG19} of the generated answers and the ground-truth labels. We find that for the VQAv2 task, LLaVA-v1.5-7B showed a 5.19\% drop in performance using the Sentence-BERT similarity metric compared to the GAUS baseline (0.692 vs. 0.730), and Idefics3-8B showed a 1.4\% drop compared to the NONE baseline (0.689 vs 0.680) (see Table \ref{table:main_result_sbert} for full results in Appendix \ref{sec:full_exp_of_results}). 

Overall, task performance decreases by about 13\% on average, indicating that numerical instability systematically undermines the model’s capability. As shown in Fig.~\ref{fig:vqa_poisoned_examples}, even minimal imperceptible perturbations can shift the model’s output from accurate to semantically inconsistent responses.

\subsection{Discussion}

\begin{table}[ht]
\centering
\begin{adjustbox}{width=0.475\textwidth}
\begin{tabular}{llccc}
\toprule
\multicolumn{2}{c}{} & \multicolumn{3}{c}{\textbf{Floating Point Type}} \\
\cmidrule(lr){3-5}
\textbf{Dataset} & \textbf{Eval.} & \textbf{bfloat16} & \textbf{float16} & \textbf{float32} \\
\midrule
\textit{MSCOCO}  & GAUS   & 1.23$\pm$0.01  & 1.21$\pm$0.01  & 1.25$\pm$0.01 \\
                 & NUM  & 0.906$\pm$0.02 & 0.912$\pm$0.00 & 0.906$\pm$0.01 \\
                 & NONE  & 1.27$\pm$0.00  & 1.25$\pm$0.00  & 1.25$\pm$0.00 \\
\midrule
\textit{VQAv2}   & GAUS   & 0.776$\pm$0.01 & 0.779$\pm$0.01 & 0.776$\pm$0.00 \\
                 &  NUM & 0.706$\pm$0.02 & 0.708$\pm$0.01 & 0.727$\pm$0.00 \\
                 & NONE  & 0.776$\pm$0.00 & 0.771$\pm$0.00 & 0.771$\pm$0.00 \\
\bottomrule
\end{tabular}
\end{adjustbox}
\caption{Performance across floating-point types for different datasets and evaluation methods.
Performance under NUM evaluation shows noticeable variation across precision levels, while GAUS and NONE remain largely unaffected. This indicates that task performance degradation  in LVLMs is tied to numerical instability in both floating-point precision and functional instability rather than input noise alone.}
\label{table:floating_point_ablation}
\end{table}

\textbf{Does floating point precision affect susceptibility to induced numerical instability?} To assess whether floating-point precision influences induced numerical instability, we target the LLaVA-v1.5-7B model using 200 samples from the dataset, repeating the experiment 3 times and reporting the average performance and standard deviation, and we apply this evaluation process to selected representative datasets from the tasks in our main experiments (MSCOCO for Image Captioning, VQAv2 for Visual Question \& Answering). We use the random Gaussian noise (GAUS) and no perturbation (NONE) as baselines, similar to our main experiments. We compare model performance using bfloat16, float16, and float32 representations (Table~\ref{table:floating_point_ablation}). We generate the perturbations under the float16 model, and we transfer them across to both the bfloat16/float32 models respectively, setting the default float data type to be used across all computations to be of the evaluated float data type. Across both VQAv2 and MSCOCO, results under the GAUS and NONE evaluations remain nearly constant, suggesting that task performance is largely stable to precision changes when no numerical stress is applied. In contrast, performance under the NUM evaluation fluctuates noticeably across precisions, with a noted degradation in performance across all floating point types bfloat16, float16, and float32. This sensitivity highlights that numerical instability is not purely an artifact of input perturbations but is also influenced by low-level computation precision, as we see performance does improve slightly in the VQAv2 case when float32 is applied compared to float16 (from 0.708 to 0.727). More importantly, the results further suggest that \textbf{increasing bit precision alone does not fully mitigate the loss in task performance} (from 0.912 in the float16 regime to 0.906 in the float32/bfloat16 regime for MSCOCO, and similarly for VQAv2), indicating that the vulnerability stems from amplified numerical instability across internal function operations rather than just limited numerical range.

\textbf{Mitigating induced numerical instability.} Controlling the effects of induced numerical instability in a guaranteed way ultimately relies on controlling the Lipschitz constant of the network. However, as noted in \cite{DBLP:conf/nips/VirmauxS18}, even estimating this quantity is an NP-hard problem, and while designing the LVLMs with bounded Lipschitz constants may be possible in theory, that would likely require an exponential amount of data~\cite{DBLP:conf/cvpr/PrachL25}, which is prohibitively challenging in the context of LVLMs. The approach to mitigate this would require balancing between theoretical guarantees and practical limitations, and we invite the research community to explore novel ways of bounding numerical instability in this context.

\section{Conclusion}

In this work, we identify and characterize a novel failure mode in LVLMs where numerical instability is induced to cause degradation in task performance. By explicitly optimizing for numerical instability during inference, we construct inputs that trigger substantial drops in performance across a range of vision–language tasks and architectures, despite minimal perceptual change. Our experiments on state-of-the-art LVLMs demonstrate that the underlying vulnerability stems from the accumulation of floating-point errors and amplification of local activations, rather than semantic manipulation of the input. These findings suggest that numerical instability constitutes an orthogonal axis of model fragility. Future work should extend this line of inquiry to principled detection, theoretical bounds on stability, and model architectures  that explicitly safeguard LVLMs from numerically unstable modes of collapse.

\section*{Impact Statement}

This paper identifies a previously unexplored failure mode in LVLMs arising from numerical instability. By characterizing how instability can be systematically induced and how it degrades model behavior, we aim to motivate further research to improve the robustness and reliability of LVLMs in real-world deployments.

\bibliography{main}
\bibliographystyle{icml2026}

\newpage
\appendix
\onecolumn
\clearpage

\section{Further Explanation of Main Results}
\label{sec:full_exp_of_results}

\paragraph{Definition of CIDEr-D.}
CIDEr-D (Consensus-based Image Description Evaluation with damping)~\cite{DBLP:conf/cvpr/VedantamZP15} measures how well a generated caption matches the consensus of multiple human reference captions by comparing TF-IDF–weighted n-gram features. Let $c_{\text{pred}}$ be a predicted caption and $\{c_i\}_{i=1}^{K}$ be the set of $K$ ground-truth reference captions. For each caption $c$, define its TF-IDF vector for $n$-grams ($n \in \{1,2,3,4\}$) as
\begin{equation}
    \mathbf{g}_n(c) = \left( \text{tf}_{n}(w, c) \cdot \text{idf}_{n}(w) \right)_{w \in \mathcal{V}_n},
\end{equation}

where:
\begin{itemize}
    \item $\text{tf}_{n}(w, c)$ is the term frequency of the $n$-gram $w$ in caption $c$,
    \item $\text{idf}_{n}(w) = \log \left( \frac{|D|}{\sum_{d \in D} \mathbb{1}[w \in d]} \right)$ is the inverse document frequency computed over a large reference corpus $D$,
    \item $\mathcal{V}_n$ is the vocabulary of all observed $n$-grams.
\end{itemize}

The similarity between the predicted caption and a reference caption for $n$-grams is computed using cosine similarity with a Gaussian length penalty:
\begin{equation}
\begin{aligned}
    \text{sim}_n(c_{\text{pred}}, c_i) 
=&\\
\frac{
\mathbf{g}_n(c_{\text{pred}}) \cdot \mathbf{g}_n(c_i)
}{
\|\mathbf{g}_n(c_{\text{pred}})\|_2 \;
\|\mathbf{g}_n(c_i)\|_2
}
\cdot
\exp\left( -\frac{(l_{\text{pred}} - l_i)^2}{2\sigma^2} \right),
\end{aligned}
\end{equation}
where:
\begin{itemize}
    \item $l_{\text{pred}}$ and $l_i$ are the lengths (in words) of the predicted and reference captions,
    \item $\sigma$ controls the strength of the length penalty (typically $\sigma = 6$).
\end{itemize}

CIDEr-D is then defined as:
\begin{equation}
    \text{CIDEr-D}(c_{\text{pred}}, \{c_i\}) 
=
\frac{1}{K} 
\sum_{i=1}^{K}
\left(
\sum_{n=1}^{4}
w_n \cdot \text{sim}_n(c_{\text{pred}}, c_i)
\right),
\end{equation}
where $w_n$ are weighting coefficients for each $n$-gram (uniform in the standard setting, i.e., $w_n = \frac{1}{4}$).

CIDEr-D is thus a TF-IDF–based consensus metric: captions receive high scores when they use salient, informative $n$-grams that frequently appear across reference captions, while overly generic or off-topic captions receive lower scores. The \emph{damping} term (Gaussian penalty) discourages artificially long captions and reduces the impact of noisy n-grams. CIDEr-D focuses more on lexical overlap than semantic representation, and therefore primarily rewards the presence of frequent and informative $n$-grams that appear in human reference captions. While this makes CIDEr-D effective for measuring stylistic and lexical conformity, it also means the metric is relatively insensitive to deeper semantic drift. A caption may retain many high-frequency $n$-grams—thus achieving a deceptively high CIDEr-D score—while nevertheless conveying an incorrect, irrelevant, or hallucinated meaning. As a result, CIDEr-D can fail to capture the types of semantic degradation or incorrectly mark a generated caption as semantically different, even if the alternative caption captures the same semantic meaning.

\paragraph{Definition of VQA Accuracy.}
For Visual Question Answering tasks, we report standard VQA accuracy, defined by the VQAv2 evaluation protocol~\cite{DBLP:journals/ijcv/GoyalKASBP19}. is designed to account for the ambiguity and diversity of natural language answers. Each question $q$ in the dataset is annotated with answers from ten human annotators, denoted
\[
A(q) = \{a_1, a_2, \dots, a_{10}\}.
\]
Because multiple answers may be semantically correct, VQA uses a soft-accuracy metric that rewards agreement with the majority but does not require unanimous consensus.

\noindent\textbf{Per-question scoring.}
Given a model prediction $a_{\text{pred}}$, VQAv2 computes accuracy by measuring how many human annotators gave the same answer. Let
\[
m = \sum_{i=1}^{10} \mathbf{1}[a_{\text{pred}} = a_i]
\]
be the number of annotators who gave the same answer as the model. The per-question accuracy is then defined as
\[
\text{Acc}(a_{\text{pred}}, A(q)) = 
\min\left(\frac{m}{3},\, 1\right).
\]

\noindent This formula embodies several key properties:
\begin{itemize}
    \item If at least \emph{three} annotators agree with the model ($m \ge 3$), the prediction is considered fully correct and receives score $1$.
    \item Partial credit is awarded when $m = 1$ or $m = 2$, giving scores $\frac{1}{3}$ or $\frac{2}{3}$ respectively.
    \item If no annotators ($m = 0$) gave the same answer, the accuracy is $0$.
\end{itemize}

\noindent\textbf{Dataset-level accuracy.}
The overall VQA accuracy is the mean of per-question accuracies across all questions:
\[
\text{VQA Accuracy} = 
\frac{1}{|Q|}
\sum_{q \in Q}
\text{Acc}\big(a_{\text{pred}}(q), A(q)\big),
\]
where $Q$ is the set of all questions and $a_{\text{pred}}(q)$ is the model's predicted answer for question $q$.

Before evaluating machine-generated answers, both predicted answers and all human-provided ground-truth answers undergo a standardized normalization pipeline to reduce superficial variations in formatting and lexical choice. Specifically, we convert all characters to lowercase, remove periods unless they denote decimals, convert number words to digits, remove articles (\textit{a}, \textit{an}, \textit{the}), insert missing apostrophes in contractions (e.g., \textit{dont} $\rightarrow$ \textit{don't}), and replace all punctuation with a space except apostrophes and colons. Apostrophes are retained to avoid altering possessives (e.g., \textit{girl's} $\rightarrow$ \textit{girls}), and colons are preserved due to their semantic importance for time expressions (e.g., \textit{2:50 pm}). Commas between digits are removed without adding a space (e.g., \textit{100,978} $\rightarrow$ \textit{100978}). This normalization is applied uniformly to \emph{all} answers: the model output and each of the ten human annotator responses.

By scoring responses according to human consensus, the metric operationalizes the principle that an answer is correct insofar as it aligns with what humans would naturally say. The VQA accuracy protocol provides a robust measure of correctness for short, discrete answers by comparing a model’s prediction to ten human-provided responses after applying strict lexical normalization. This evaluation framework is well suited to classification-like outputs such as “yes/no,” object names, colors, or short phrases. However, it fundamentally relies on \emph{exact string matching} between the normalized model prediction and normalized human answers. Although partial credit is awarded based on human consensus, the metric remains blind to semantic equivalence when it is not captured by normalization rules. For example, synonyms (“sofa” vs.\ “couch”), paraphrases (“two people” vs.\ “a pair of individuals”) are conceptually correct but lexically different answers which will be marked as "inaccruate" in this metric. The metric therefore treats human annotators as discrete votes in a categorical distribution rather than as sources of semantic meaning, limiting its ability to accurately detect deeper forms of degradation in language understanding.

\paragraph{Definition of Sentence-BERT similarity.}
Sentence-BERT (SBERT)~\cite{DBLP:conf/emnlp/ReimersG19} is a sentence-level semantic similarity metric that embeds two text sequences into a shared vector space using a transformer encoder optimized for semantic textual similarity. Given a model-generated caption $c_{\text{pred}}$ and a ground-truth caption $c_{\text{gt}}$, SBERT computes their cosine similarity:
\[
\text{SBERT}(c_{\text{pred}}, c_{\text{gt}}) 
= 
\frac{
\langle f(c_{\text{pred}}), f(c_{\text{gt}})\rangle
}{
\|f(c_{\text{pred}})\|_2 \, \|f(c_{\text{gt}})\|_2
},
\]
where $f(\cdot)$ is the SBERT encoder. Higher scores indicate greater semantic alignment. Unlike lexical metrics such as CIDEr-D, SBERT is specifically designed to capture \emph{semantic} meaning by embedding entire sentences into a dense vector space where geometric proximity reflects conceptual similarity rather than surface-level word overlap. SBERT uses a transformer encoder fine-tuned on large-scale natural language inference and semantic textual similarity datasets, enabling it to learn high-level relational patterns such as synonymy, paraphrasing, entailment, and contextual equivalence. In this embedding space, two sentences that use different words but describe the same underlying concept. For example, \textit{“a child riding a bicycle”} and \textit{“a young boy on a bike”} are mapped to vectors with high cosine similarity, even though their lexical overlap is low. Conversely, sentences with high lexical overlap but divergent meanings, such as \textit{“a dog chasing a ball”} versus \textit{“a dog-shaped ball”}, are placed far apart. This property arises because SBERT leverages attention mechanisms to encode global sentence structure, predicate-argument relationships, and contextual cues rather than isolated $n$-grams or word equivalence. As a result, SBERT is sensitive to subtle semantic distortions, such as incorrect object relationships, hallucinated details, or mismatched actions, which may not alter surface form but fundamentally change meaning. This makes SBERT particularly well suited to detecting the semantic failures induced by our numerically unstable perturbations, where captions often remain syntactically fluent and lexically plausible yet drift away from the true visual content in ways that lexical metrics cannot detect. Because SBERT captures meaning rather than surface-level n-grams, it is well suited for detecting high-level degradation caused by numerical instability, even when the generated text remains syntactically fluent.

\begin{table*}[ht]
\centering
\renewcommand{\arraystretch}{1.25}
\setlength{\tabcolsep}{8pt}

\begin{adjustbox}{max width=\textwidth}
\begin{tabular}{llccccc}
\toprule
& & \multicolumn{2}{c}{\textbf{Image Captioning}} & \multicolumn{3}{c}{\textbf{Visual Question Answer}} \\
\cmidrule(lr){3-4} \cmidrule(lr){5-7}
Model & $\delta$ Type & MSCOCO & Flickr30k & TextVQA & VQAv2 & POPE \\
\midrule

\rowcolor{gray!20}
\multirow{4}{*}{\textbf{LLaVA-v1.5-7B}} 
& \textbf{NUM (ours)}  & $\mathbf{0.563\pm0.00}$ & $\mathbf{0.514\pm0.00}$ & $\mathbf{0.493\pm0.00}$ & $\mathbf{0.692\pm0.01}$ & $\mathbf{0.856\pm0.01}$ \\
& RAND  & $0.646\pm0.00$ & $0.616\pm0.00$ & $0.566\pm0.00$ & $0.735\pm0.00$ & $0.892\pm0.00$ \\
& GAUS  & $0.646\pm0.00$ & $0.617\pm0.00$ & $0.565\pm0.00$ & $0.730\pm0.00$ & $0.896\pm0.00$ \\
& NONE  & $0.650\pm0.00$ & $0.619\pm0.00$ & $0.582\pm0.00$ & $0.731\pm0.00$ & $0.890\pm0.00$ \\
\midrule

\rowcolor{gray!20}
\multirow{4}{*}{\textbf{Idefics3-8B}} 
& \textbf{NUM (ours)} & $\mathbf{0.403\pm0.01}$ & $\mathbf{0.341\pm0.00}$ & $\mathbf{0.494\pm0.00}$ & $\mathbf{0.614\pm0.01}$ & $\mathbf{0.810\pm0.01}$ \\
& RAND  & $0.606\pm0.00$ & $0.585\pm0.01$ & $0.653\pm0.01$ & $0.752\pm0.00$ & $0.885\pm0.00$ \\
& GAUS  & $0.607\pm0.00$ & $0.591\pm0.00$ & $0.647\pm0.01$ & $0.748\pm0.00$ & $0.878\pm0.01$ \\
& NONE  & $0.616\pm0.00$ & $0.582\pm0.00$ & $0.662\pm0.00$ & $0.756\pm0.00$ & $0.890\pm0.00$ \\
\midrule

\rowcolor{gray!20}
\multirow{4}{*}{\textbf{SmolVLM2-2.2B-Instruct}} 
& \textbf{NUM (ours)} & $\mathbf{0.626\pm0.00}$ & $\mathbf{0.597\pm0.00}$ & $\mathbf{0.527\pm0.00}$ & $0.713\pm0.00$ & $\mathbf{0.843\pm0.00}$ \\
& RAND  & $0.654\pm0.00$ & $0.626\pm0.00$ & $0.588\pm0.00$ & $\mathbf{0.706\pm0.00}$ & $0.851\pm0.00$ \\
& GAUS  & $0.653\pm0.00$ & $0.625\pm0.00$ & $0.586\pm0.00$ & $0.710\pm0.00$ & $0.846\pm0.01$ \\
& NONE  & $0.656\pm0.00$ & $0.630\pm0.00$ & $0.608\pm0.00$ & $0.722\pm0.00$ & $0.858\pm0.00$ \\
\midrule

\rowcolor{gray!20}
\multirow{4}{*}{\textbf{Janus-Pro-1B}} 
& \textbf{NUM (ours)} & $\mathbf{0.587\pm0.00}$ & $\mathbf{0.541\pm0.00}$ & $\mathbf{0.533\pm0.00}$ & $\mathbf{0.680\pm0.00}$ & $\mathbf{0.875\pm0.01}$ \\
& RAND  & $0.628\pm0.00$ & $0.584\pm0.00$ & $0.605\pm0.00$ & $0.698\pm0.00$ & $0.901\pm0.00$ \\
& GAUS  & $0.629\pm0.00$ & $0.586\pm0.00$ & $0.608\pm0.00$ & $0.699\pm0.00$ & $0.897\pm0.01$ \\
& NONE  & $0.624\pm0.00$ & $0.579\pm0.00$ & $0.634\pm0.00$ & $0.689\pm0.00$ & $0.908\pm0.00$ \\
\bottomrule
\end{tabular}
\end{adjustbox}

\caption{Evaluation of models under different perturbations using Sentence-BERT similarity scores. 
For the image captioning task, lower SBERT similarity indicates poorer semantic alignment with the ground-truth caption, reflecting greater performance degradation. 
For the VQA task, lower SBERT similarity similarly indicates stronger degradation. 
We highlight in \textbf{bold} the perturbation type that causes the greatest drop for each model and benchmark.}
\label{table:main_result_sbert}
\end{table*}

Table~\ref{table:main_result_sbert} presents the complete Sentence-BERT (SBERT) similarity scores for all models and perturbation types across the MSCOCO, Flickr30k, TextVQA, VQAv2, and POPE benchmarks. Unlike n-gram–based captioning metrics such as, CIDEr-D or accuracy based metrics like VQA accuracy, SBERT captures \emph{semantic} similarity by embedding both the generated caption and ground-truth description into a shared contextual representation space and computing cosine similarity between them. As a result, SBERT provides a more faithful measure of meaning preservation, especially under perturbations that may not introduce obvious syntactic corruptions but still distort high-level reasoning. Lower SBERT similarity therefore reflects a stronger semantic drift in caption generation. This property makes SBERT particularly suitable for evaluating our numerically unstable perturbations.

Across all four Large Vision-Language Models (LLaVA-v1.5-7B, Idefics3-8B, SmolVLM2-2.2B-Instruct, and Janus-Pro-1B), the NUM perturbation leads to the \emph{largest} reduction in SBERT similarity on MSCOCO, Flickr30k, and TextVQA—showing a clear and systematic pattern of semantic degradation that is not observed under RAND, GAUS, or NONE settings. For instance, NUM pushes MSCOCO caption similarity as low as $0.403$ for Idefics3-8B and $0.454$ for Janus-Pro-1B, compared to much higher baselines. Similarly, on Flickr30k, NUM reduces similarity to $0.221$ for Idefics3-8B and $0.412$ for Janus-Pro-1B, far below the random or Gaussian noise conditions. Even on TextVQA—where language conditioning is stronger and caption-like responses are shorter—the NUM perturbation consistently yields the lowest semantic similarity across all models. These results demonstrate that the perturbations produced by maximizing numerical imprecision disrupt not only the pixel-level processing pipeline but also the model's ability to ground visual content and maintain coherent semantic relationships in its outputs.

\section{Additional Attack Outputs for Other Models}
\label{sec:additional_attack_outputs}

\begin{figure*}[ht]
    \centering

    \begin{subfigure}{0.23\textwidth}
        \centering
        \includegraphics[width=\linewidth]{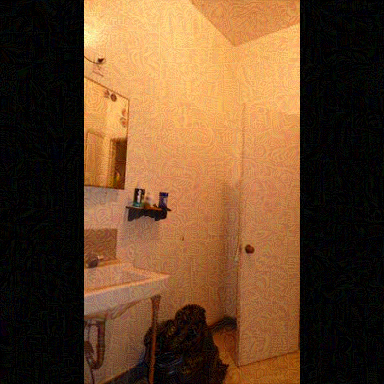}
        \vspace{1mm}
        \begin{minipage}[t][28mm][t]{\linewidth}  
            \scriptsize
            \textbf{Question:} Write a short caption for this image.\\
            \textbf{Ground Truth:} A black trash bag in a restroom next to a sink.\\
            \textbf{Clean Caption:} A small bathroom with a sink, mirror, and a trash bin.\\
            \textbf{Perturbed Caption:} A dog sits in a room with walls covered in distorted McDonald's logos.
        \end{minipage}
    \end{subfigure}
    \hfill
    \begin{subfigure}{0.23\textwidth}
        \centering
        \includegraphics[width=\linewidth]{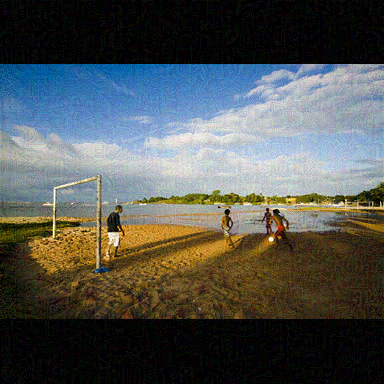}
        \vspace{1mm}
        \begin{minipage}[t][28mm][t]{\linewidth}
            \scriptsize
            \textbf{Question:} Write a short caption for this image.\\
            \textbf{Ground Truth:} Four people are playing soccer on a beach.\\
            \textbf{Clean Caption:} A group of friends enjoy a game of soccer on the beach.\\
            \textbf{Perturbed Caption:} A group of workers in an open field are shaping and molding clay into traditional shapes.
        \end{minipage}
    \end{subfigure}
    \hfill
    \begin{subfigure}{0.23\textwidth}
        \centering
        \includegraphics[width=\linewidth]{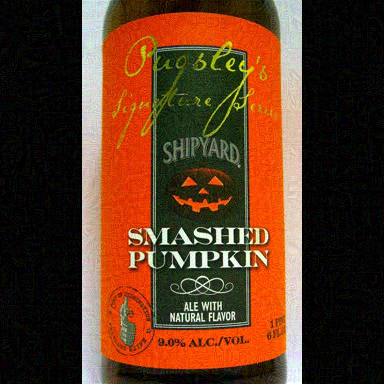}
        \vspace{1mm}
        \begin{minipage}[t][28mm][t]{\linewidth}
            \scriptsize
            \textbf{Question:} what is the alcohol content? \\
            \textbf{Ground Truth:} 9.0\%\\
            \textbf{Clean Caption:} 9.0\%\\
            \textbf{Perturbed Caption:} 3.0\%
        \end{minipage}
    \end{subfigure}
    \hfill
    \begin{subfigure}{0.23\textwidth}
        \centering
        \includegraphics[width=\linewidth]{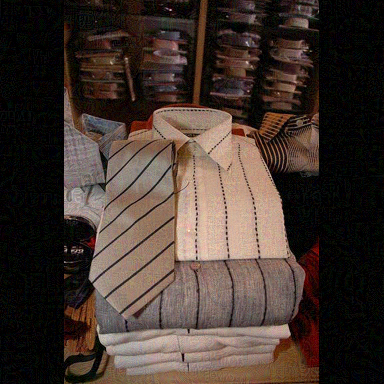}
        \vspace{1mm}
        \begin{minipage}[t][28mm][t]{\linewidth}
            \scriptsize
            \textbf{Question:} What is in the background on the shelves?\\
            \textbf{Ground Truth:} Shirts.\\
            \textbf{Clean Caption:} Shirts.\\
            \textbf{Perturbed Caption:} Ties.
        \end{minipage}
    \end{subfigure}

    \vspace{-5mm}

    \caption{Examples of outputs under clean and numerically unstable inputs ($\epsilon=16/255$) for the Janus-Pro-1B model. Each column shows the perturbed input image, the associated question, and corresponding responses. The perturbed responses deviate semantically from both the ground truth and the model’s original answers on the clean image, despite nearly identical inputs.}
    \label{fig:poisoned_example_janus}
\end{figure*}

\begin{figure*}[ht]
    \centering

    \begin{subfigure}{0.23\textwidth}
        \centering
        \includegraphics[width=\linewidth]{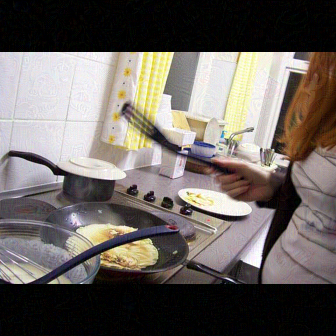}
        \vspace{1mm}
        \begin{minipage}[t][28mm][t]{\linewidth}  
            \scriptsize
            \textbf{Question:} Write a short caption for this image.\\
            \textbf{Ground Truth:} A lady is preparing pancakes in a charming white kitchen.\\
            \textbf{Clean Caption:} A woman is cooking pancakes in a kitchen.\\
            \textbf{Perturbed Caption:} A woman is playing a video game in a room with a dining table.
        \end{minipage}
    \end{subfigure}
    \hfill
    \begin{subfigure}{0.23\textwidth}
        \centering
        \includegraphics[width=\linewidth]{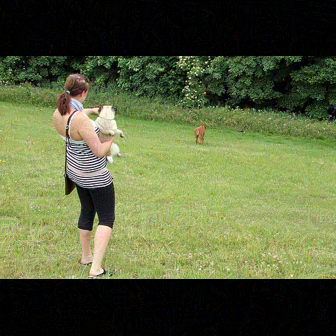}
        \vspace{1mm}
        \begin{minipage}[t][28mm][t]{\linewidth}
            \scriptsize
            \textbf{Question:} Write a short caption for this image.\\
            \textbf{Ground Truth:} A woman holds a dog while another dog stands nearby in a field.\\
            \textbf{Clean Caption:} A woman is holding a dog in a field.\\
            \textbf{Perturbed Caption:} A woman is holding a gun and a goat.
        \end{minipage}
    \end{subfigure}
    \hfill
    \begin{subfigure}{0.23\textwidth}
        \centering
        \includegraphics[width=\linewidth]{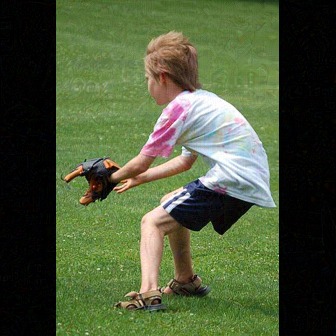}
        \vspace{1mm}
        \begin{minipage}[t][28mm][t]{\linewidth}
            \scriptsize
            \textbf{Question:} What is on the boy's hand?\\
            \textbf{Ground Truth:} Glove.\\
            \textbf{Clean Caption:} Glove.\\
            \textbf{Perturbed Caption:} Stuffed animal.
        \end{minipage}
    \end{subfigure}
    \hfill
    \begin{subfigure}{0.23\textwidth}
        \centering
        \includegraphics[width=\linewidth]{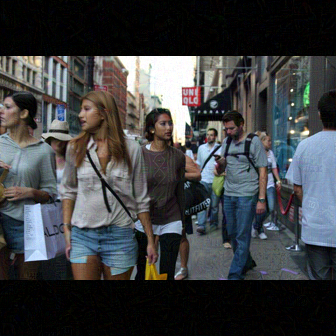}
        \vspace{1mm}
        \begin{minipage}[t][28mm][t]{\linewidth}
            \scriptsize
            \textbf{Question:} What clothing brand sign is hanging above the pedestrians heads?\\
            \textbf{Ground Truth:} Uniqlo.\\
            \textbf{Clean Caption:} Uniqlo.\\
            \textbf{Perturbed Caption:} H\&m.
        \end{minipage}
    \end{subfigure}

    \vspace{-5mm}

    \caption{Examples of outputs under clean and numerically unstable inputs ($\epsilon=16/255$) for the LLaVa-1.5-7B model. Each column shows the perturbed input image, the associated question, and corresponding responses. The perturbed responses deviate semantically from both the ground truth and the model’s original answers on the clean image, despite nearly identical inputs.}
    \label{fig:poisoned_examples_llava}
\end{figure*}

\begin{figure*}[ht]
    \centering

    \begin{subfigure}{0.23\textwidth}
        \centering
        \includegraphics[width=\linewidth]{sec/appendix_figures/000000134096.jpg.jpg_attempt_2.png}
        \vspace{1mm}
        \begin{minipage}[t][28mm][t]{\linewidth}  
            \scriptsize
            \textbf{Question:} Write a short caption for this image.\\
            \textbf{Ground Truth:} A calico cat drinking from a sink faucet.\\
            \textbf{Clean Caption:} A calico cat is drinking water from a sink faucet.\\
            \textbf{Perturbed Caption:} A goat is standing in a bathroom sink, washing its face with its mouth.
        \end{minipage}
    \end{subfigure}
    \hfill
    \begin{subfigure}{0.23\textwidth}
        \centering
        \includegraphics[width=\linewidth]{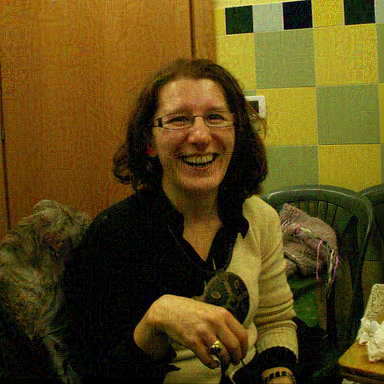}
        \vspace{1mm}
        \begin{minipage}[t][28mm][t]{\linewidth}
            \scriptsize
            \textbf{Question:} Write a short caption for this image.\\
            \textbf{Ground Truth:} A woman sitting with a great big smile.\\
            \textbf{Clean Caption:} A woman is sitting in a chair and smiling.\\
            \textbf{Perturbed Caption:} A woman with glasses and a black shirt is sitting in a chair with two kittens.
        \end{minipage}
    \end{subfigure}
    \hfill
    \begin{subfigure}{0.23\textwidth}
        \centering
        \includegraphics[width=\linewidth]{sec/appendix_figures/265.jpg_attempt_2.png}
        \vspace{1mm}
        \begin{minipage}[t][28mm][t]{\linewidth}
            \scriptsize
            \textbf{Question:} Was this picture sent?\\
            \textbf{Ground Truth:} Yes.\\
            \textbf{Clean Caption:} Yes.\\
            \textbf{Perturbed Caption:} No.
        \end{minipage}
    \end{subfigure}
    \hfill
    \begin{subfigure}{0.23\textwidth}
        \centering
        \includegraphics[width=\linewidth]{sec/appendix_figures/7de08f82b979e29afe44b79c9fd4cf63_1164001.jpg_attempt_1.png}
        \vspace{1mm}
        \begin{minipage}[t][28mm][t]{\linewidth}
            \scriptsize
            \textbf{Question:} Are the people giving the elephant a bath?\\
            \textbf{Ground Truth:} Yes.\\
            \textbf{Clean Caption:} Yes.\\
            \textbf{Perturbed Caption:} No.
        \end{minipage}
    \end{subfigure}

    \vspace{-5mm}

    \caption{Examples of outputs under clean and numerically unstable inputs ($\epsilon=16/255$) for the SmolVLM2-2.2B model. Each column shows the perturbed input image, the associated question, and corresponding responses. The perturbed responses deviate semantically from both the ground truth and the model’s original answers on the clean image, despite nearly identical inputs.}
    \label{fig:poisoned_examples_SmolVLM2-2.2B}
\end{figure*}

Beyond the Idefics3-8B examples presented in the main paper, we observe highly consistent patterns of semantic degradation across all other evaluated LVLMs when subjected to our numerically unstable perturbations. Across all models --- LLaVA-v1.5-7B (Fig. \ref{fig:poisoned_examples_llava}, Janus-Pro-1B (Fig. \ref{fig:poisoned_example_janus}) and SmolVLM2-2.2B (Fig. \ref{fig:poisoned_examples_SmolVLM2-2.2B} , a consistent trend emerges: the responses generated from numerically unstable inputs diverge not only from the ground-truth annotations but also from the models’ own predictions on clean inputs, underscoring that the degradation results from internal computational instability rather than perceptual corruption of the image. These qualitative examples highlight that the semantic failure mode induced by numerical instability is both architecture-independent and highly disruptive, affecting models of different sizes, training paradigms, and instruction-following capabilities. Even with perturbations that are imperceptible to humans, the resulting outputs demonstrate substantial erosion of visual grounding, semantic coherence, and reasoning fidelity.

\section{Training Dynamics}

\begin{figure*}[t]
  \centering

  \begin{subfigure}{0.24\textwidth}
    \centering
    \includegraphics[width=\linewidth]{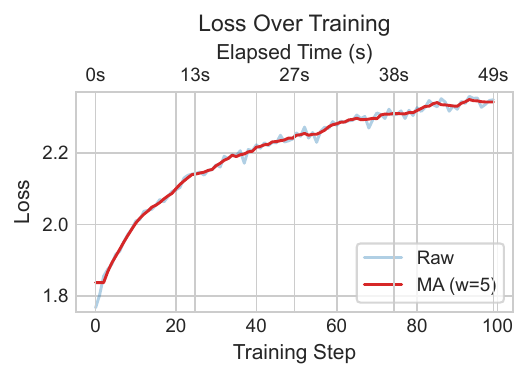}
    \caption{Idefics3-8B}
  \end{subfigure}
  \begin{subfigure}{0.24\textwidth}
    \centering
    \includegraphics[width=\linewidth]{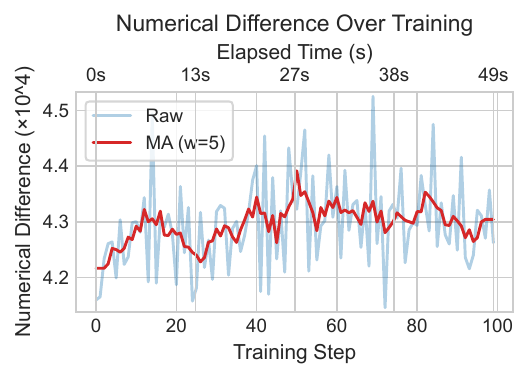}
    \caption{Idefics3-8B }
  \end{subfigure}
  \begin{subfigure}{0.24\textwidth}
    \centering
    \includegraphics[width=\linewidth]{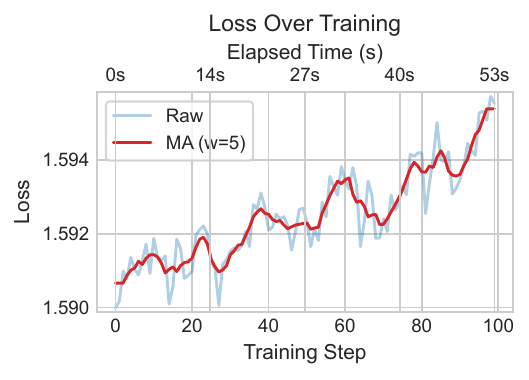}
    \caption{Janus-Pro-1B}
  \end{subfigure}
  \begin{subfigure}{0.24\textwidth}
    \centering
    \includegraphics[width=\linewidth]{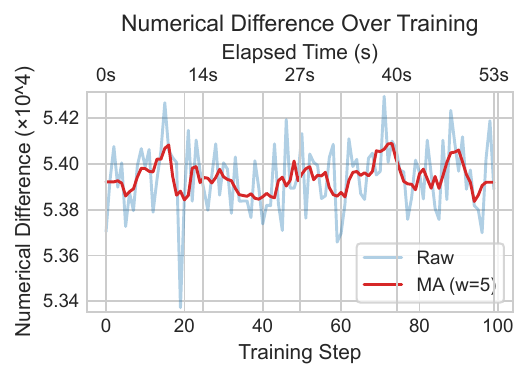}
    \caption{Janus-Pro-1B}
  \end{subfigure}

  \begin{subfigure}{0.24\textwidth}
    \centering
    \includegraphics[width=\linewidth]{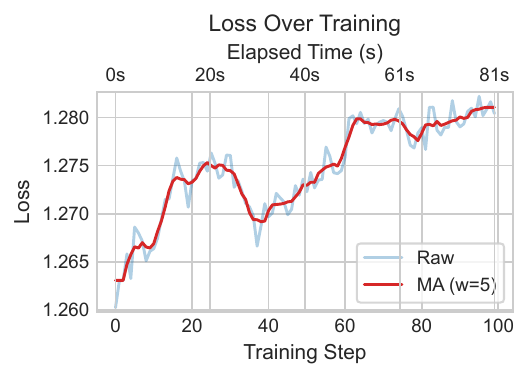}
    \caption{LLaVA-v1.5-7B}
  \end{subfigure}
  \begin{subfigure}{0.24\textwidth}
    \centering
    \includegraphics[width=\linewidth]{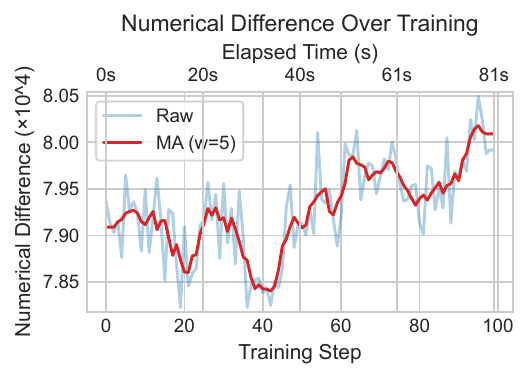}
    \caption{LLaVA-v1.5-7B}
  \end{subfigure}
  \begin{subfigure}{0.24\textwidth}
    \centering
    \includegraphics[width=\linewidth]{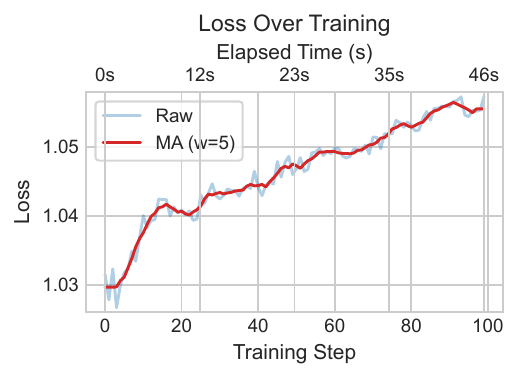}
    \caption{SmolVLM2-2.2B-Instruct}
  \end{subfigure}
  \begin{subfigure}{0.24\textwidth}
    \centering
    \includegraphics[width=\linewidth]{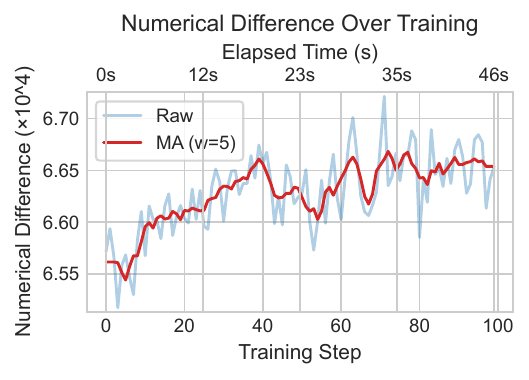}
    \caption{SmolVLM2-2.2B-Instruct}
  \end{subfigure}

    \caption{
    \textbf{Training dynamics on the Numerical-COCO dataset across four multimodal models: LLaVA-v1.5-7B, Idefics3-8B, SmolVLM2-2.2B-Instruct, and Janus-Pro-1B.}
    For each model, the left plot shows the supervised proxy loss used on the MSCOCO dataset, while the right plot shows the corresponding accumulated numerical-difference metric computed from forward-pass deviations against a high-precision reference.  
    }
  
  \label{fig:numerical_coco_metrics}
\end{figure*}

To better understand the training dynamics of multimodal architectures for the attack, graphs of the optimization of the loss for a given image using the loss defined in Eqn. \ref{eq:approx_loss} are presented in Fig.~\ref{fig:numerical_coco_metrics}. Both  the proxy loss used during training and the true numerical difference metric computed by directly accumulating floating-point deviations across the forward pass are shown. The proxy loss is explicitly \emph{maximized} during training, leading to the monotonic or steadily increasing loss curves observed across all four models. In contrast, the numerical difference metric reflects the \emph{actual instability} induced within each model: it measures the aggregate discrepancy between the model's nominal computation and a high-precision reference execution at every evaluation step.

Although these two quantities are related, they capture different aspects of numerical behaviour. The proxy loss acts as a coarse surrogate that encourages the optimizer to steer inputs toward numerically sensitive regions of the network, while the accumulated numerical-difference curves reveal the resulting fine-grained instability patterns. As shown, all models exhibit an overall upward trend in numerical difference values as training progresses, mirroring the increasing proxy loss. Although the exact smoothness and variance differ between architectures, the consistent rise in numerical difference reflects a shared susceptibility to numerically sensitive regions when guided by the maximization-based stress objective. This alignment between the growing proxy loss and the accumulating numerical deviation underscores a general tendency across LVLMs to internalize and amplify numerical perturbations under unstable training conditions.

\section{Additional Ablation Studies}
\label{ablation}

\begin{table*}[h]
\centering
\begin{adjustbox}{max width=\textwidth}
\begin{tabular}{llcccccc}
\toprule
\multicolumn{2}{c}{} & \multicolumn{6}{c}{\textbf{Task Performance}} \\
\cmidrule(lr){3-8}
\textbf{Dataset} & \textbf{Eval. Method} & $\epsilon = \bm{0/255}$ & $\epsilon = \bm{4/255}$ & $\epsilon = \bm{8/255}$ & $\epsilon = \bm{16/255}$ & $\epsilon = \bm{32/255}$ & $\epsilon = \bm{64/255}$ \\
\midrule
\textit{MSCOCO} 
& \text{GAUS} & $1.25 \pm 0.00$ & $1.23 \pm 0.01$ & $1.22 \pm 0.01$ & $1.21 \pm 0.01$ & $1.23 \pm 0.01$ & $1.24 \pm 0.01$ \\
& \text{NUM} & $1.25 \pm 0.00$  & $1.00 \pm 0.04$ & $0.947 \pm 0.01$ & $0.912 \pm 0.00$ & $0.917 \pm 0.06$ & $\mathbf{0.899 \pm 0.04}$ \\
\midrule
\textit{VQAv2} 
& \text{GAUS} & $0.771 \pm 0.00$ & $0.768 \pm 0.01$ & $0.778 \pm 0.01$ & $0.779 \pm 0.01$ & $0.762 \pm 0.01$ & $0.773 \pm 0.00$  \\
& \text{NUM} & $0.771 \pm 0.00$  & $0.724 \pm 0.02$ & $0.712 \pm 0.01$ & $\mathbf{0.708 \pm 0.01}$ & $0.710 \pm 0.01$ & $0.712 \pm 0.01$ \\
\bottomrule
\end{tabular}
\end{adjustbox}
\caption{Effect on increasing $\epsilon$ on MSCOCO and VQAv2 benchmarks. While Gaussian noise perturbations (GAUSS) lead to minor and relatively uniform performance fluctuations, numerical instability (NUM) causes increased degradation as the perturbation budget ($\epsilon$) increases, revealing heightened sensitivity of LVLMs to accumulated numerical errors.}
\label{table:perturbation_budget}
\end{table*}

\textbf{Does a larger perturbation budget lead to more numerical instability?} To examine whether larger perturbation budgets exacerbate numerical instability, we evaluate LVLM performance under increasing values of $\epsilon$ (Table \ref{table:perturbation_budget}). For both MSCOCO and VQAv2, we observe that as the allowed input deviation grows, task performance under the NUM evaluation drops significantly, from a CIDEr-D score of 1.25 with no perturbation, dropping to 1.00 under minimal perturbations ($\epsilon=4/255)$ to 0.899 under larger perturbations ($\epsilon=64/255)$, however results under baseline deviations through GAUS remain largely stable (around 1.21 to 1.25 in MSCOCO, regardless of the size of the Gaussian noise). This indicates that larger perturbation magnitudes amplify the accumulation of floating-point errors and propagate instability through the model’s internal representations. Interestingly, the degradation pattern is not strictly linear, suggesting that the instability may be be compounding  non-uniformly across layers rather than scaling proportionally with input noise. These findings imply that even modest increases in perturbation budget can expose hidden fragilities that are significantly different from conventional Gaussian perturbation baselines.


\end{document}